\PassOptionsToPackage{table,xcdraw}{xcolor}
\documentclass[runningheads]{llncs}
\usepackage[T1]{fontenc}
\usepackage{tikz}
\usepackage{amssymb}
\usepackage{pifont}
\usepackage{graphicx}
\usepackage{hhline}
\usepackage{amsmath}
\usepackage[colorlinks=true, linkcolor=blue, citecolor=blue, urlcolor=blue]{hyperref}
\usepackage{subcaption}
\usepackage{array}
\usepackage{comment}
\usepackage{booktabs}
\usepackage{tcolorbox}
\usepackage{enumitem}
\usepackage[table,xcdraw]{xcolor}
\usepackage{multirow}
\begin{document}
\title{
Concept-based Rubrics Improve LLM Formative Assessment and Data Synthesis 
}
\titlerunning{Rubric-Improved LLM Assessment Ability}
%

\author{
  Yuchen Wei\inst{1} \and
  Matthew Beckman\inst{2} \and
  Dennis Pearl\inst{2} \and
  Rebecca J. Passonneau\inst{1}
}

\institute{
  Department of Computer Science and Engineering,\\
  Pennsylvania State University, State College, PA 16801, USA\\
  \email{\{yxw5769, rjp49\}@psu.edu}
  \and
  Department of Statistics,\\
  Pennsylvania State University, State College, PA 16801, USA\\
  \email{\{mdb268, dkp13\}@psu.edu}
}

\maketitle              
\begin{abstract}

Formative assessment in STEM topics aims to promote student learning by identifying students' current understanding, thus targeting how to promote further learning. Previous studies suggest that the 
assessment performance of current generative large language models (LLMs) on constructed responses to open-ended questions is significantly lower than that of supervised classifiers trained on 
high-quality labeled data. 
However, we demonstrate that 
concept-based rubrics can significantly enhance LLM performance, which narrows 
the gap between LLMs as off-the shelf assessment tools, and smaller supervised models, which need large amounts of training data. 
For datasets where concept-based rubrics allow LLMs to achieve strong performance, 
we show that the concept-based rubrics help the same LLMs generate high quality synthetic data for training lightweight, high-performance supervised models. 
Our experiments span diverse STEM student response datasets
with labels of varying quality, including a new real-world dataset that contains some AI-assisted responses, which introduces additional considerations.

\keywords{Automated Assessment \and Formative Assessment \and Large Language Models \and Pre-trained Language Models \and Rubrics \and Data Synthesis}

\end{abstract}
%
%
%

\section{Introduction}

Formative assessment aims to provide students and instructors with feedback regarding individual students' current learning, thus helping both learners and educators target the next steps to advance a student's learning. While numerous studies have shown the importance of timeliness in provision of feedback \cite{hattie&timperley07,fisherEtAl25,boud&molloy13}, it can be difficult to provide this in large introductory STEM classes. Our work investigates automated formative assessment of constructed response questions in STEM subjects, to support timeliness. There has been much previous work on supervised methods that incorporate pre-trained language models (PLMs) in student response assessment (e.g., \cite{wang2019inject,camus2020investigating}), which can achieve expert-level performance (e.g. \cite{AsRRN}). Supervised methods, however, incur the high cost of creating sufficient amounts of training data, and performance does not generalize well across STEM domains. On the flip side, generative large language models (LLMs) require no training data, 
generalize well across STEM domains, and have the potential to generate informative comments.
However, they have not previously achieved the same accuracy as supervised methods \cite{llmshortanswerscoring,kortemeyer24}. 
Here we show how use of concept-based rubrics designed for human use narrows the performance gap between the two methods. The rubrics also improve the quality of LLM-generated synthetic data for training supervised models. 
We also provide speculative results on feedback comments provided by LLMs. 

\begin{table}[t!]
    \centering
    \small
    \begin{tabular}{l | l|r|r|r|r|c}
        \multicolumn{1}{c|}{Dataset}
        & \multicolumn{1}{c|}{Domain}
        & \multicolumn{1}{c|}{Vintage}
        & \multicolumn{1}{c|}{Students}        
        & \multicolumn{1}{c|}{Labelers}
        & \multicolumn{1}{c|}{Format}
        & \multicolumn{1}{c}{Rubrics} \\\hline
    Beetle & electronics & 2000s & UG & NLP researchers & standal. & n \\
    SciEntsBank & many STEM & 2000s & 3rd-6th & NLP researchers & standal. & n \\
    ASAP-SAS & STEM + ELA & 2000s & 7/8/10 & Trained raters & mix & y \\
    ISTUDIO & statistics & 2010s & UG & Stat Ed Experts & scenario & y \\
    CLASSIFIES & statistics & 2020s & UG & Trained stat TAs & scenario & y \\
    \end{tabular} 
    \vspace{.1in}
    \caption{\small Five datasets and their domains (including English Language Arts, ELA); decade of collection (Vintage); student level; expertise of labelers; whether the questions are standalone or accompany a scenario; inclusion of rubrics.}
    \label{tab:datasets}
    \vspace{-.2in}
\end{table}

Our experiments rely on a range of publicly available datasets 
in diverse STEM subjects, listed in Table~\ref{tab:datasets}. They differ in time of origin, relevant to the likelihood of AI-generated responses; 
question format, such as standalone questions versus scenarios with multiple questions; 
presence of a rubric, and rubric type. 
The diversity of datasets enhances the generalizability of our results.

We first show, consistent with previous work, that supervised PLMs have high performance on our datasets. We then confirm that performance of GPT 4o-mini, a state-of-the-art LLM, is lower than supervised methods. A disadvantage of LLMs is that maximizing performance typically depends on trial-and-error prompt engineering~\cite{liuEtAl-llmprompting-23}. However, we find a dramatic performance increase when pre-existing content-based rubrics are used as prompts, 
outperforming prompting with examples. 
The same rubrics improve the quality of synthetic data generated by GPT 4o-mini, as demonstrated by experiments where we use the synthetic data to train a lightweight, PLM-based classifier. 

\section{Related Work}

Currently, two main approaches are widely used in research on automated answer assessment: supervised training of end-to-end neural networks, often involving fine-tuning of PLMs, and in-context learning with large language models. Among  supervised approaches, SFRN \cite{SFRN} employs relational networks \cite{relationnetwork} 
to learn the relationship between an input question, student answer, and reference answers; BERT \cite{bert} was used as the initial encoder. AsRRN \cite{AsRRN} is a more advanced relation network using recurrence \cite{recurrentrelationnetwork} along with a contrastive loss function \cite{contrastivelearning}. 
Fine-tuned PLMs like RoBERTa \cite{roberta} have also been shown to achieve strong performance \cite{longanswergrading,thakkar2021finetuning}. 
Ega{\~n}a et al. \cite{ASAG_annotationstratigies} employ RoBERTa-MNLI, a RoBERTa model fine-tuned on the MNLI dataset \cite{mnli}, to explore few-shot strategies on the SemEval dataset \cite{semeval2013}.

With the emergence of large language models (LLMs), evaluating their effectiveness in student answer assessment tasks has become a popular research area. Several studies have explored this topic. For example, one study \cite{llmshortanswerscoring} compared the performance of fine-tuned BERT models with in-context learning using few-shot examples from LLMs, including GPT \cite{GPT} and LLaMA \cite{llama}, on short-answer datasets. Another study \cite{longanswergrading} evaluated the performance of LLMs against RoBERTa-MNLI on the RiceChem dataset \cite{longanswergrading}. Additionally, a recent study \cite{gpt-4-semEval} demonstrated that 
incorporating XML \cite{XML} with
answer samples as input to GPT-4 can achieve strong performance on the SemEval dataset.


Previous research has explored using synthetic data generated by LLMs to support text classifier training. For example, one study demonstrated that LLMs can generate synthetic data from diverse domains to improve the generalization ability of NLI models \cite{synthetic_data_for_NLI}. Another work \cite{Datagenerationfortextclassification} investigated the effectiveness of mixing LLM-generated synthetic data with a few real-world samples in many common text classification tasks. Some research  highlighted that highly subjective text classification tasks, such as humor detection, gain little benefit from LLM-generated data \cite{synthetic_data_generation_with_llm}. In our work, we explore methods for training supervised PLMs for the answer assessment task, which is a subjective judgment task, using data generated by LLMs, such as student responses and scoring annotations, to replicate the scoring capability of LLMs in a lightweight PLM. Unlike previous work that required real-world student responses to distill the LLM’s capabilities for answer assessment \cite{chatgptforexplainableautomatedstudentanswerassessment}, our approach can replicate LLM capabilities using purely LLM-generated data without human data effort. This is achieved by increasing data diversity and re-annotating synthetic samples. 

\begin{table}[t!]
    \centering
    \small
\begin{tabular}{l|r|r|r|r|r|r}
         \multicolumn{1}{c|}{Dataset (\# of Tokens)} 
         & \multicolumn{1}{c|}{Mean}
         & \multicolumn{1}{c|}{Median}
         & \multicolumn{1}{c|}{Min.}
         & \multicolumn{1}{c|}{Max.}
         & \multicolumn{1}{c|}{\# Questions}
         & \multicolumn{1}{c}{\# Responses} \\\hline
    Beetle               & 10.9 & 10.0 & 1 &  84 &  56 & 5,653\\
    SciEntsBank          & 14.6 & 13.0 & 2 & 131 & 197 &  11,135\\
    ASAP SAS             & 51.9 & 48.0 & 1 & 392 &  10 &  25,683 \\
    ISTUDIO              & 42.0 & 35.0 & 1 & 567 &   6 &  6,532\\
    CLASSIFIES           & 73.1 & 56.0 & 1 & 690 &   8 &  5,819 \\   
\end{tabular}
    \vspace{.1in}
    \caption{\small Descriptive statistics including number of questions, answer length in tokens (based on BERT \cite{bert} tokenization), and number of student responses.}
    \label{tab:dataset_num_tokens}
\end{table}

\section{Datasets}
\label{sec:datasets}

The experiments were carried out on five datasets to comprehensively evaluate methods across diverse assessment tasks: Beetle and SciEntsBank \cite{semeval2013}, ISTUDIO \cite{AsRRN}, ASAP-SAS \cite{shermis2014} and a new dataset, CLASSIFIES. Table \ref{tab:datasets} presents some key differences among the datasets. Table \ref{tab:dataset_num_tokens} provides the average answer lengths, number of questions, and number of responses.


The \textbf{BEETLE and SciEntsBank} datasets \cite{semeval2013} were developed to evaluate student responses to short-answer questions, aiming to support the development of automated assessment methods. 
These two datasets were originally developed with multiple label sets (2-way, 3-way and 5-way) and for three types of machine learning challenges: unseen answers, unseen questions, and unseen domains. 
Student responses were annotated with one of five labels: \emph{Correct}, \emph{Partially Correct}, \emph{Incomplete}, \emph{Contradictory}, \emph{Irrelevant}, and \emph{Non-domain}. In this work, 
for comparability with our other datasets, we adopt a 3-way classification scheme consisting of \emph{Correct}, \emph{Partially Correct}, and \emph{Incorrect}, where the latter includes \emph{Contradictory}, \emph{Irrelevant}, and \emph{Non-domain}.


\textbf{ISTUDIO}  (for Introductory Statistics Transfer of Understanding and Discernment Outcomes \cite{ISTDUIO}) was designed for assessing 
statistical reasoning abilities in undergraduates with an introductory level statistics background from all over the U.S.  There are three scenarios, each accompanied by two questions.  Responses are classified as
\emph{Essentially Correct}, \emph{Partially Correct}, or \emph{Incorrect}. 
Detailed, rubrics were prepared for human assessors, along with example responses for each label. The rubrics define specific conceptual components required for a correct answer, outline different constraints on these components for partially correct answers, and identify possible cases of incorrect answers.

\textbf{ASAP} (Automated Student Assessment Prize) was created as part of a public competition to evaluate the performance of automated scoring systems for constructed essay responses \cite{ASAP,shermis2014}. The dataset includes 
responses from six U.S. states: three on the east coast, two in the midwest, and one on the west coast. 
The responses are labeled using holistic scoring rubrics. Scores range from 0–2 or 0–3 depending on the task prompt. In this work, we use a 3-way subset of ASAP, focusing on 6 task prompts with scores ranging from 0–2. 

\textbf{CLASSIFIES} (Common Language Assessment in Studying Statistics with Instructional Feedback and Increased Enrollment Scalability) is a new dataset with undergraduate responses to statistics questions utilized in classrooms from six U.S. universities.\footnote{Download link will be provided if paper is accepted.} Student responses were annotated by trained undergraduates using a 3-way labeling scheme: \emph{Correct}, \emph{Partially Correct}, or \emph{Incorrect}. A subset was ready for use at the time of this research consisting of student responses from 
two universities. 
Each question includes a question text, a model solution, a sample response, and a detailed component-based rubric, which defines different levels of correctness, with each level corresponding to specific components of the model solution. 
\section{Methods}

To establish target performance, we train PLMs on our datasets as described in the first subsection below. The next subsection discusses three ways of using LLMs: prompting LLMs for student response assessment; data synthesis for training PLMs; generation of feedback comments.



\subsection{Supervised PLMs}

To train lightweight neural networks on our assessment tasks, we fine-tuned pre-trained language models (PLMs), including RoBERTa and Longformer, which are widely used in answer assessment \cite{longanswergrading,ASAG_annotationstratigies,thakkar2021finetuning} and other text classification tasks \cite{synthetic_data_generation_with_llm,Datagenerationfortextclassification}. For shorter text inputs (total input $\leq$ 512 tokens), we used RoBERTa-large \cite{roberta}. For longer text input ($>$ 512 tokens), we used Longformer-large \cite{longformer}, a variant of RoBERTa-large designed for longer sequences. Regarding the training settings, we trained the PLMs for 10 epochs with an initial learning rate of 2e-5. For datasets with a predefined validation set (CLASSIFIES, ISTUDIO), we used their respective validation sets. 
For Beetle, SciEntsBank and ASAP, we allocated 10\% of the training data for validation.

\subsection{Large Language Models (LLMs) for Multiple Tasks}
\label{subsec:LLM-methods}
For all three tasks we explore with LLMs, we use
GPT4o-Mini due to its superior performance and cost-effectiveness.

\begin{figure}[t!]
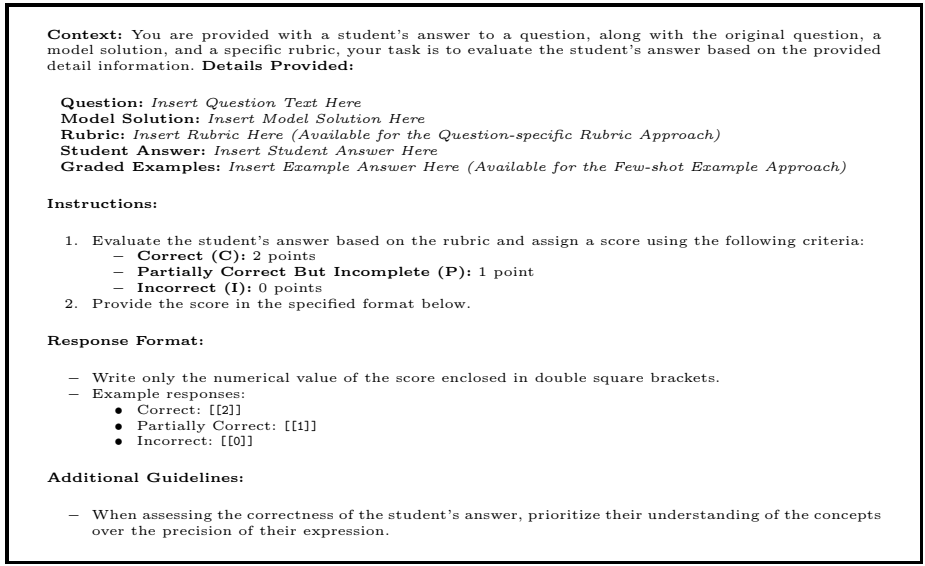

\begin{tiny}
\begin{tcolorbox}[colframe=black, colback=white, sharp corners, boxrule=0.5mm]
\textbf{Context:} You are provided with a student's answer to a question, along with the original question, a model solution, and a specific rubric, your task is to evaluate the student's answer based on the provided detail information.
\textbf{Details Provided:}
\begin{itemize}[left=0pt, label={}]
    \item \textbf{Question:} \textit{Insert Question Text Here}
    \item \textbf{Model Solution:} \textit{Insert Model Solution Here}
    \item \textbf{Rubric:} \textit{Insert Rubric Here (Available for the Question-specific Rubric Approach)}
    \item \textbf{Student Answer:} \textit{Insert Student Answer Here}
    \item \textbf{Graded Examples:} \textit{Insert Example Answer Here (Available for the Few-shot Example Approach)}
\end{itemize}

\textbf{Instructions:}
\begin{enumerate}[label=\arabic*.]
    \item Evaluate the student's answer based on the rubric and assign a score using the following criteria:
    \begin{itemize}
        \item \textbf{Correct (C):} 2 points
        \item \textbf{Partially Correct But Incomplete (P):} 1 point
        \item \textbf{Incorrect (I):} 0 points
    \end{itemize}
    \item Provide the score in the specified format below.
\end{enumerate}

\textbf{Response Format:}
\begin{itemize}
    \item Write only the numerical value of the score enclosed in double square brackets.
    \item Example responses:
    \begin{itemize}
        \item Correct: \texttt{[[2]]}
        \item Partially Correct: \texttt{[[1]]}
        \item Incorrect: \texttt{[[0]]}
    \end{itemize}
\end{itemize}

\textbf{Additional Guidelines:}
\begin{itemize}
    \item When assessing the correctness of the student's answer, prioritize their understanding of the concepts over the precision of their expression.
\end{itemize}

\end{tcolorbox}
\end{tiny}
    \vspace{.1in}
\caption{\small Prompt template for evaluating GPT 4o-mini performance.}
\label{fig:GPT-evaluation-template}
\end{figure}

\textbf{Student Response Assessment.}
LLM performance on many tasks can often be improved by using prompts that demonstrate a task, or explain how to carry it out, known as in-context learning \cite{dong-etal-2024-survey}.
We compared the effectiveness of incorporating no examples, one or more examples, and prompting with rubrics instead of examples. The type of template we used is shown in Figure~\ref{fig:GPT-evaluation-template}. 
We experimented with from 0 to 5 examples for each of the three output class labels (see section~\ref{sec:datasets}). All examples were randomly selected from the training data.

\textbf{LLMs to Synthesize New Data.}
We compared three methods of data synthesis: (1) using the LLM to label the training set of actual student responses, (2) using the LLM to generate a fully synthetic training set, meaning responses and their labels, and (3) quality-enhancements to response synthesis. For the first method, we prompted GPT 4o-mini with the question, a model solution (a correct example response), 
the rubric, and a student response. 
For the second method, we used
the LLM to generate both a response text and its label, a technique explored in various other text classification tasks \cite{synthetic_data_generation_with_llm,xshot,Datagenerationfortextclassification}. 
To synthesize responses, the prompt included the question, the model solution, the rubric, and the target label. For the second method, we noticed that synthesized responses seemed to sometimes not match the label in the prompt. To address the issue of possible mismatches between the response and the input label, we relabeled the synthesized responses with GPT4o-mini. Responses also lacked diversity, even with a high-temperature setting (1.3). Therefore, we experimented with enhancements to our second data synthesis method. 

To enhance the diversity of generated responses, we combined two strategies. First, we prompted the LLM to generate a list of elements that occur in the concept-based rubric and to generate multiple case statements corresponding to subsets of these elements, and their correctness labels. Then we re-prompted the LLM to generate a response for each case statement. The second strategy was to vary lengths of responses by including a target length in the prompt, randomly selected from a uniform distribution ranging from 5 to 128 words.

\textbf{Explanatory Feedback:}
Recent research has used 
LLMs to create more explainable and student-centered assessments \cite{validityofautomaticallygeneratedfeedback,chatgptforexplainableautomatedstudentanswerassessment}. In this work, we focus on the LLM's ability to generate feedback that explains its assessment label. Specifically, for datasets with effective question-specific rubrics, such as ISTUDIO and CLASSIFIES, we instruct the LLM to not only assess student answers but also explain the rationale for the assessment, drawing on the content of the student responses and the rubrics. 


Feedback comments generated by GPT 4o-mini were manually annotated on the three following dimensions:
\begin{itemize}
    \item \textbf{Label Correctness}: In cases where the human and LLM label disagree, which label seems more correct with respect to the rubric?  (\textit{Human} or \textit{LLM}).
    \item \textbf{Explainability} Does the LLM explanation reasonably 
    justify its predicted assessment based on the student's answer and the rubric? (\textit{Yes} or \textit{No}).
    \item \textbf{Subjectivity}: 
    Does the LLM explanation favor subjective factors such as clarity or stylistic factors over alignment with the rubric?
    (\textit{Yes} or \textit{No}).
\end{itemize}
To assess \textit{Label Correctness}, we randomly selected 50 examples each from GPT4o-mini's incorrect predictions for ISTUDIO and CLASSIFIES.  Literature on scaffolding student learning indicates that feedback is most useful when a students' responses are \emph{Partially Correct}, to build upon their current understanding \cite{bellandEtAl_scaffolding_2017}. Thus, to assess \textit{Explainability}, we randomly selected 50 student responses per dataset from GPT4o-mini's correct predictions of \emph{Partially Correct}.

\section{Experiments, Results and Discussion}
\label{sec:results}

We present results from four types of experiments comparing supervised PLMs for automated assessment with unsupervised LLMs. First we establish the performance benchmark PLMs achieve as a target for LLMs to reach or surpass. Then we show that while rubrics do not benefit PLMs, concept-based rubrics greatly boost the performance of LLMs. Our third set of experiments investigates the potential of LLMs prompted with rubrics for synthesis of training data, to reduce the cost of PLMs. Finally, we present preliminary results on the potential for LLM assessment to include explanatory feedback.

\subsection{Supervised PLMs} 
\label{subsec:PLMs}
Here we compare assessment accuracy of supervised PLMs across different types of inputs when trained on human-generated, human-labeled data. As expected, PLMs can achieve high accuracy.

\begin{table}[t!]
\centering
\begin{tabular}{l|l|c|c|c}
\hline
\textbf{Dataset} & \textbf{Model} & \textbf{Input Components} & \textbf{Accuracy (\%)} & \textbf{F1 (\%)} \\ \hline
\multirow{3}{*}{\textbf{Beetle 3-way}} 
  & \multirow{3}{*}{\textbf{PLM-R}} & \textbf{Q} + \textbf{A} + \textbf{M} + \textbf{R} & 83.21\% & 82.93\% \\ \cline{3-5}
  &  & \textbf{Q} + \textbf{A} + \textbf{M}               & \textbf{86.04}\% & \textbf{85.95\%} \\ \cline{3-5}
  &  & \textbf{A} + \textbf{M}                           & 85.35\% & 85.24\% \\ \hline\hline
\multirow{3}{*}{\textbf{SciEntsBank 3-way}}
  & \multirow{3}{*}{\textbf{PLM-R}} & \textbf{Q} + \textbf{A} + \textbf{M} + \textbf{R} & 81.00\% & 81.63\% \\ \cline{3-5}
  &  & \textbf{Q} + \textbf{A} + \textbf{M}               & 81.00\% & 81.25\% \\ \cline{3-5}
  &  & \textbf{A} + \textbf{M}                           & \textbf{81.05\%} & \textbf{81.84\%} \\ \hline\hline
\multirow{3}{*}{\textbf{ISTUDIO}}
  & \multirow{3}{*}{\textbf{PLM-L}} & \textbf{Q} + \textbf{A} + \textbf{M} + \textbf{R} & \textbf{86.59\%} & \textbf{86.62\%} \\ \cline{3-5}
  &  & \textbf{Q} + \textbf{A} + \textbf{M}               & 85.42\% & 85.58\% \\ \cline{3-5}
  &  & \textbf{A} + \textbf{M}                           & 84.84\% & 84.76\% \\ \hline\hline
\multirow{3}{*}{\textbf{CLASSIFIES}}
  & \multirow{3}{*}{\textbf{PLM-L}} & \textbf{Q} + \textbf{A} + \textbf{M} + \textbf{R} & 83.01\% & 83.06\% \\ \cline{3-5}
  &  & \textbf{Q} + \textbf{A} + \textbf{M}               & 82.56\% & 82.66\% \\ \cline{3-5}
  &  & \textbf{A} + \textbf{M}                           & \textbf{83.41\%} & \textbf{83.31\%} \\ \hline\hline
\multirow{3}{*}{\textbf{ASAP}}
  & \multirow{3}{*}{\textbf{PLM-L}} & \textbf{Q} + \textbf{A} + \textbf{M} + \textbf{R} & \textbf{73.41\%} & \textbf{73.44\%} \\ \cline{3-5}
  &  & \textbf{Q} + \textbf{A} + \textbf{M}               & 73.12\% & 73.31\% \\ \cline{3-5}
  &  & \textbf{A} + \textbf{M}                           & 72.53\% & 72.52\% \\ 
\hline
\end{tabular}
\vspace{1em}
\caption{\small Performance of supervised PLMs (\textbf{PLM-R} for RoBERTa and \textbf{PLM-L} for Longformer) with different inputs selected from the \textbf{Q}{uestion}, \textbf{A}{nswer}, \textbf{M}{odel Solution}, and \textbf{R}{ubric}.}
\label{tab:fully_supervised_methods}
\end{table}

We trained the PLMs with different subsets of input from the
(\textbf{Q}{uestion}, Student \textbf{A}{nswer}, \textbf{M}{odel answer}, or \textbf{R}{ubric}). Table \ref{tab:fully_supervised_methods} illustrates the results across different datasets. The performance of fine-tuned PLMs is impressive, particularly when compared with the performance of GPT4o-Mini discussed later in section \ref{subsec:ICL}. 
Comparison of the rows with the rubric (\textbf{Q} + \textbf{A} + \textbf{M} + \textbf{R}) with other rows shows low and inconsistent benefit of the rubric 
across different datasets. This includes ISTUDIO and CLASSIFIES, whose concept-based rubrics greatly boost the performance of LLMs, as shown below. 
Beetle, SciEntsBank and ASAP have simpler, stand-alone questions in comparison to the scenario-based questions in ISTUDIO and CLASSIFIES. Beetle, ISTUDIO and CLASSIFIES represent a single domain and  have relatively few questions, in comparison to the other datasets. All these factors presumably affect performance.

\subsection{LLM In-Context Learning: Concept-based Rubrics Can Help}
\label{subsec:ICL}
\begin{figure}[t!]
    \centering
    \includegraphics[width=0.9\linewidth]{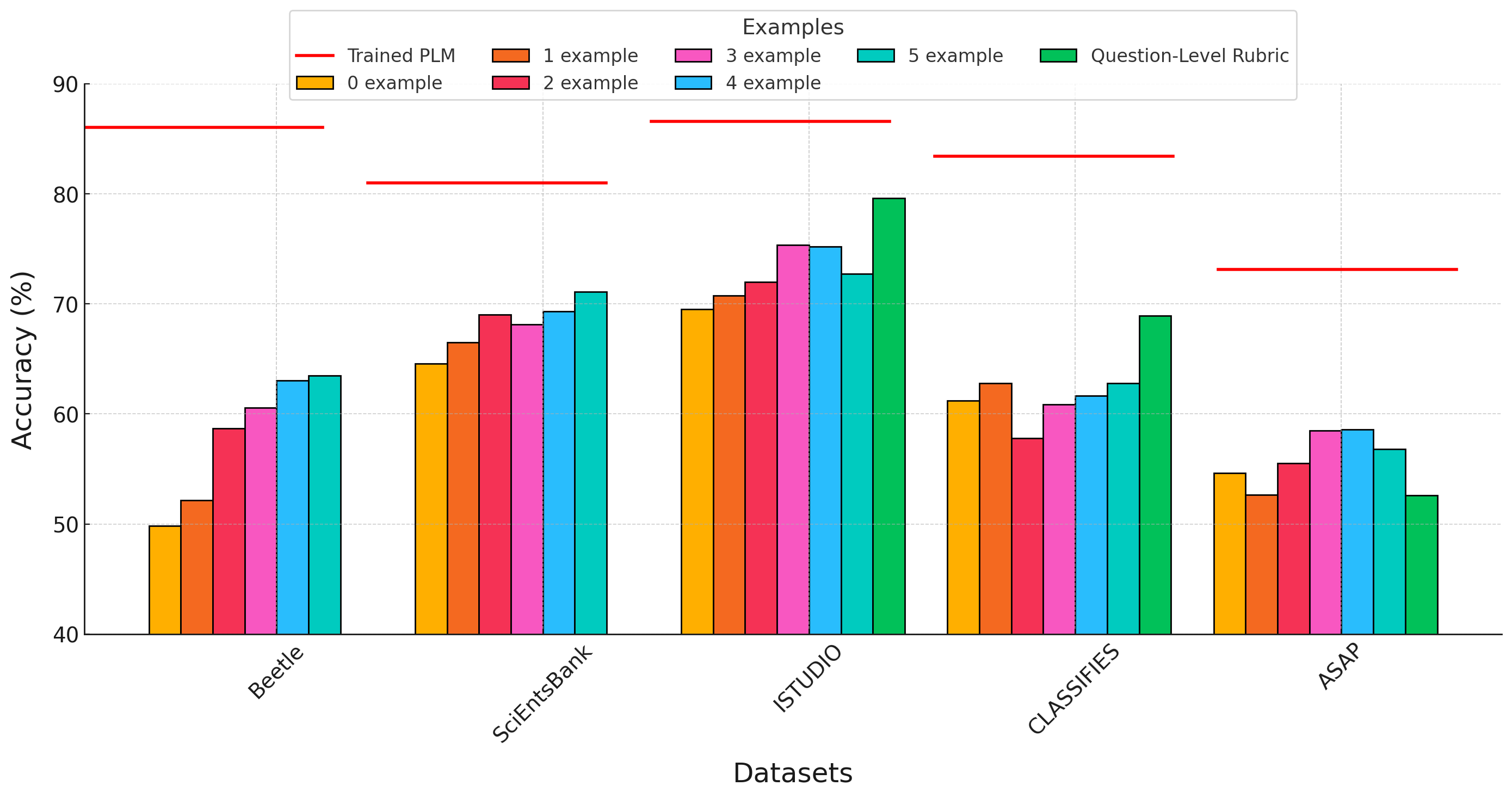}
    
    \caption{\small GPT4o-Mini performance across different datasets, incorporating different numbers of examples and the question-specific rubric. Note that \emph{number of examples} is per label per question; 
    5 examples per label corresponds to \(3 \times 5 = 15\) examples per prompt.
    }
    \label{fig:GPT4o-Mini_performance}
\end{figure}

Figure \ref{fig:GPT4o-Mini_performance} presents the results of GPT4o-Mini prompted with from 0 to 5 examples per label, or, for all but the SemEval datasets, a question-specific rubric. 
We observe that \textbf{each additional example improves accuracy}, with gains ranging from 5\% to 15\%. This result aligns with the intuition that examples provide better context for the model. For CLASSIFIES and ISTUDIO, \textbf{question-specific rubrics yield better performance compared to many examples}, resulting in approximately a 10\% improvement for ISTUDIO and an 8\% improvement for CLASSIFIES. In contrast, rubrics tie with use of two examples for worst performance on ASAP. Clearly, 
question-specific rubrics are not always helpful. 

\begin{figure}[!ht]

\begin{tcolorbox}[
colback=blue!5!white, colframe=blue!75!black, fonttitle=\bfseries]
\tiny
\textbf{ASAP Question:} \emph{How does the author organize the article? Support your response with details from the article.}

\begin{itemize}
    \item \textbf{Correct (2 points):} The student answer fulfills all the requirements of the task. The information given is text-based and relevant to the task.
    \item \textbf{Partially Correct (1 point):} The student answer fulfills some of the requirements of the task, but some information may be too general, too simplistic, or not supported by the text.
    \item \textbf{Incorrect (0 points):} The student answer does not fulfill the requirements of the task because it contains information that is inaccurate, incomplete, and/or missing altogether.
\end{itemize}

\noindent\rule{\textwidth}{0.4pt} 

\textbf{ISTUDIO Question:} \emph{Explain how you would decide whether the student has a good ear for music using this method of note identification. (Be sure to give enough detail that a classmate could easily understand your approach, and how he or she would interpret the result in the context of the problem.)}

\begin{itemize}
    \item \textbf{Correct (2 points):} The student names or paraphrases a binomial procedure, one-proportion Z-procedure, or an analogous inferential method (e.g., inferential analysis of probability, proportion, or the number of correct vs. incorrect responses).
    \item \textbf{Partially Correct (1 point):} The student:
    \begin{enumerate}[label=\alph*.]
        \item Recommends drawing a conclusion based on a point estimate (e.g., probability, proportion, or number of correctly identified notes) without naming or paraphrasing an inferential statistical method, or
        \item Names or paraphrases an inappropriate inferential method (e.g., a t-test or other unrelated or flawed statistical method).
    \end{enumerate}
    \item \textbf{Incorrect (0 points):} The student:
    \begin{enumerate}[label=\alph*.]
        \item Describes a solution based on only one note played.
        \item Bases the conclusion on something other than the probability, proportion, or number of correctly identified notes.
        \item Does not provide a coherent answer.
    \end{enumerate}
\end{itemize}

\noindent\rule{\textwidth}{0.4pt} 

\textbf{CLASSIFIES Question:} \emph{Choose one of these scenarios and identify which variable would be the explanatory variable and which would be the response variable. Provide an explanation for your choice.}

\begin{itemize}
    \item \textbf{Correct:} The student answer satisfies all the following components:
    \begin{enumerate}[label=\arabic*.]
        \item Identifies the two variables.
        \item Identifies the explanatory variable and response variable consistent with Component 1.
        \item Indicates the explanatory variable predicts (or explains) the response variable, or the response variable is dependent on the explanatory variable.
    \end{enumerate}
    \item \textbf{Partially Correct:} The student answer does not meet the criteria for "Correct" but satisfies Component 1 and/or Component 2.
    \item \textbf{Incorrect:} The student answer does not meet the rubric for "Correct" or "Partially Correct."
\end{itemize}

\end{tcolorbox}
\caption{\small Sample questions and rubrics 
for ASAP, ISTUDIO, and CLASSIFIES.}
\label{fig:dataset-rubrics}
\end{figure}

\begin{table}[t!]

    \centering
        \begin{tabular}{l|r|r}
        \hline
        \textbf{Dataset} & \multicolumn{1}{|c}{Rubric vs Model Solution} & \multicolumn{1}{|c}{Rubric vs Student Answers} \\
        \hline
        CLASSIFIES & 0.6120 & 0.4855 \\
        \hline
        ISTUDIO    & 0.5172 & 0.3368 \\
        \hline
        ASAP       & 0.2257 & 0.1028 \\
    \end{tabular}
        \vspace{.1in}
    \caption{\small Average cosine similarities of question-specific rubrics to model solutions or student responses for the three datasets with question-specific rubrics.}
    \label{tab:rubric_cosine_similarity}
\end{table}

ASAP's rubrics are question-specific 
but the rubric content does not closely relate to concepts in the questions. 
Figure \ref{fig:dataset-rubrics} illustrates an ASAP rubric which states that the criteria are to determine whether a response is "too general" or "too simplistic." In contrast, the ISTUDIO and CLASSIFIES rubrics reference specific concepts, such as "binomial procedure" or "explanatory variable." 

To quantify 
the differences between concept-based versus general rubrics, we measured text similarity between a question's rubric and its model solution, as well as between the rubric and responses 
using Sentence-BERT, an efficient and effective method to apply the cosine similarity metric to pairs of sentence embeddings \cite{sentenceBERT}. 
Table \ref{tab:rubric_cosine_similarity} presents the average cosine similarities of rubrics to model solutions or to student answers. It indicates that the average similarity of the rubric in the ASAP dataset, both with the model solution and with student answers, is lower compared to ISTUDIO and CLASSIFIES. This is evidence that concept-based, question-specific rubrics not only aid in training graders to achieve high consistency  \cite{beckman2024developing}, but can also \textbf{significantly enhance the in-context learning performance of GPT4o-mini (and similar LLMs) well above the gains achieved by incorporating numerous examples.}

\subsection{Distilling LLM Knowledge for Training PLMs}
\label{sec:data-synthesis}

\begin{table}[t!]
\small
\centering
\begin{tabular}{l|c|c|r|r}\hline
\multicolumn{1}{c}{\bf Dataset} &
    \multicolumn{1}{|c}{\bf Training Data} &
        \multicolumn{1}{|c}{\bf Model} &
            \multicolumn{1}{|c}{\bf Accuracy} &
                \multicolumn{1}{|c}{\bf F1} \\\hline
\multirow{5}{*}{ISTUDIO} 
  & ---        & GPT4oMini     & 79.63\%   & 70.27\%\\ \cline{2-5}
  & Labels\&Responses  & PLM-R & 49.77\%  & 46.52\%\\ \cline{2-5}
  & LabelsOnly  & PLM-R & 79.09\% & 78.13\%\\\cline{2-5}
  & DiversityEnhanced & PLM-R & 80.70\% & \textbf{80.33\%}\\\cline{2-5}
  & Original    & PLM-R & 86.71\%  & 86.23\%\\\hline
\multirow{5}{*}{CLASSIFIES} 
   & --- & GPT4oMini    & 68.92\% & 68.28\% \\\cline{2-5}
   &  Labels\&Responses       & PLM-R & 55.62\% & 46.32\% \\\cline{2-5}
   &  LabelsOnly & PLM-R & 67.76\% & 68.23\% \\\cline{2-5}
   & DiversityEnhanced  & PLM-R & 68.76\% & \textbf{69.05\% } \\\cline{2-5}
   & Original           & PLM-R & 83.14\% & 82.96\% \\
\end{tabular}
    \vspace{.1in}
\caption{\small Performance comparison of GPT4o-mini with RoBERTa trained on the original (costly) training data, and three types of low-cost synthesized training data: LabelsOnly, Labels\&Responses, and DiversityEnhanced synthesis.}
\label{tab:llm_synthetic_vs_llm_annotation}
\end{table}

The preceding section presented results that show large accuracy improvements for LLMs, when prompted with a concept-based rubric. Performance is still not as high as the PLMs presented in section~\ref{subsec:PLMs}, but avoids the need for large amounts of training data. Here we show that a lightweight PLM can achieve or even surpass the performance of LLMs by distilling the LLM knowledge through data synthesis. As discussed in the Methods section, we compared LLM synthesis of training labels alone, with LLM synthesis of labels and responses, and full synthesis with diversity enhancing methods. We synthesized approximately 10,000 answers for each of ISTUDIO and CLASSIFIES, using the question, the model solution, the rubric, and the target label in the prompts. 

\begin{figure}[h]
    \centering
    \includegraphics[width=0.6\linewidth]{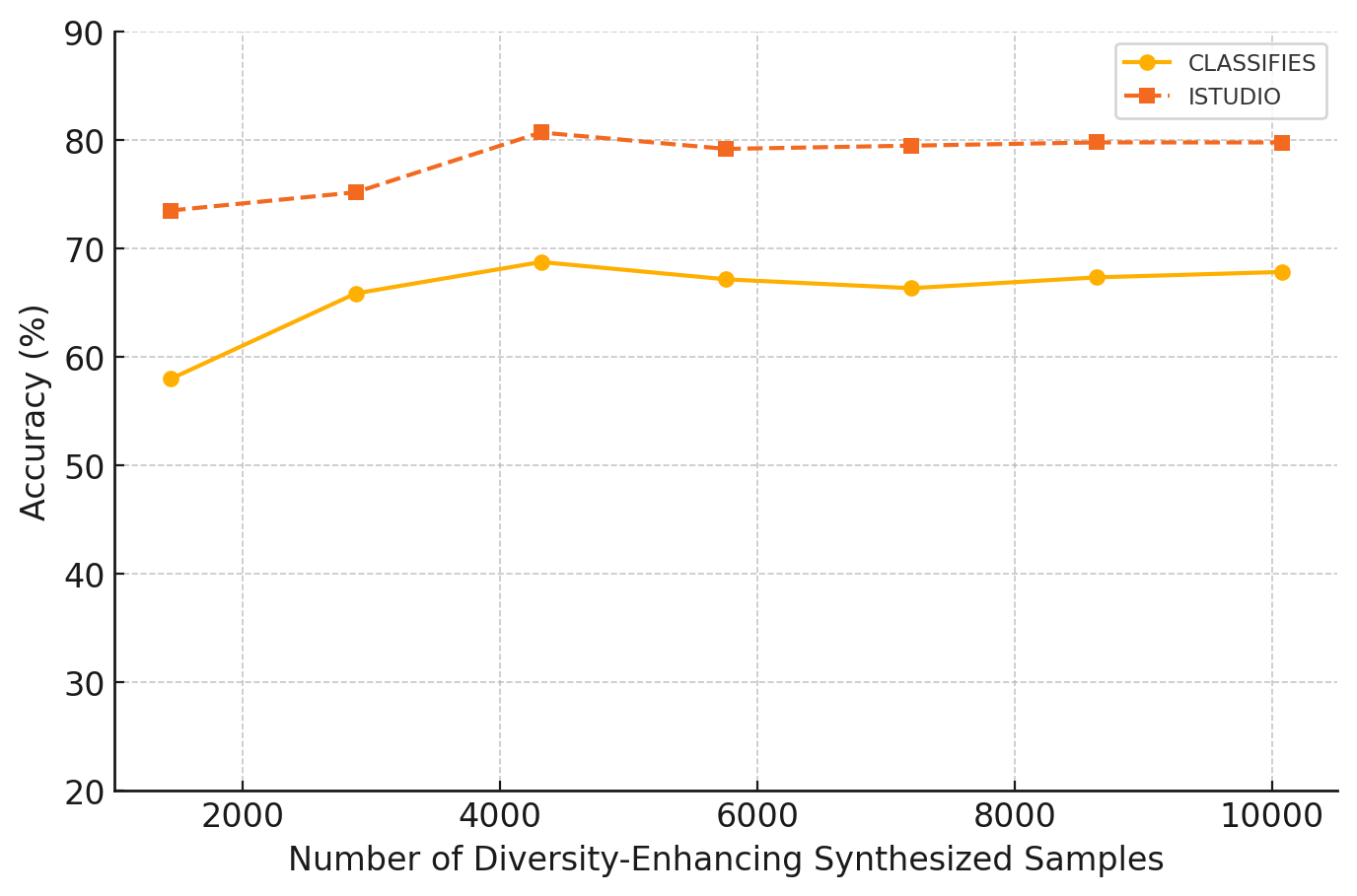}
    
    \caption{\small 
    Sensitivity of PLM accuracy to varying amounts of DiversityEnhancing synthesized samples on CLASSIFIES and ISTUDIO.
    }

    \label{fig:NumberOfSamples_Vs_Accuracy}
\end{figure}

Table \ref{tab:llm_synthetic_vs_llm_annotation} presents a comparison of the three data synthesis methods using GPT4o-mini. Using the LLM to generate labelled responses as the training set (Labels\&Responses) is ineffective for fully replicating the LLM's judgment criteria, leading to significantly lower accuracy compared to the LLM itself. However, we observe that with LLM annotations on human answers (LabelsOnly), our lightweight PLMs can 
achieve assessment accuracy comparable to the LLM. The results for the DiversityEnhancing synthesis demonstrate the effectiveness of our re-annotation and diversity-increasing strategies. After applying these improvements, RoBERTa-large trained on the synthetic data closely matches that of GPT4oMini itself. \textbf{This approach 
fully distills the capabilities of the LLM to a lightweight model, avoiding the costly human data effort.} Further, a sensitivity analysis shown in Figure \ref{fig:NumberOfSamples_Vs_Accuracy} shows that accuracy of the supervised PLM peaks with approximately 3,000 to 4,000 DiversityEnhancing training samples (reaching a performance level near that of the GPT-4o mini). 

Because the CLASSIFIES dataset was collected later than ISTUDIO, and at a time when students had access to ChatGPT, we also looked into the issue of what proportion of responses in each dataset appeared to be from an AI model like ChatGPT using GPTZero, a classifier to detect AI-generated content \cite{gptzero}. 
Table \ref{tab:gptzero_detection} shows that CLASSIFIES exhibits a significantly higher rate of AI-generated responses than ISTUDIO. This suggests that the synthetic answers generated by the Labels\&Responses method are more similar to the original dataset in CLASSIFIES, contributing to its better performance (55.62\%) compared to ISTUDIO (49.77\%). ISTUDIO has a small percentage of false positives, reflecting some uncertainty in GPTZero results.

\begin{table}[t!]
\centering
\begin{tabular}{l | c | r | r} \hline
\multicolumn{1}{c}{Dataset} & 
    \multicolumn{1}{|c}{Sample Source} & 
        \multicolumn{1}{|c}{\# of Samples} & 
            \multicolumn{1}{|c}{GPTZero Rate} \\ \hline
\multirow{2}{*}{CLASSIFIES}  & Original & 400 & 28.8\% \\ \cline{2-4}
                              & GPT4oMini Generation & 400 &  74.5\%\\ \hline
\multirow{2}{*}{ISTUDIO}     & Original & 400 & 5.8\%  \\ \cline{2-4}
                              & GPT4oMini Generation & 400 &  95.3\% \\ \hline
\end{tabular}
    \vspace{.1in}
\caption{\small GPTZero detection rates of AI-generated content for CLASSIFIES and ISTUDIO, based on 400 random selected samples from the  original training data and GPT4oMini generated data.}
\label{tab:gptzero_detection}
\end{table}

\begin{table}[t!]
\small
\centering
\begin{tabular}{l | c | c | r | r}\hline
\multicolumn{1}{c}{Dataset} &
    \multicolumn{1}{|c}{Training Data} &
        \multicolumn{1}{|c}{Method} &
            \multicolumn{1}{|c}{Accuracy} &
                \multicolumn{1}{|c}{F1} \\\hline
\multirow{4}{*}{ISTUDIO} 
    & ---          & GPT4o        & 78.10\% & 71.70\% \\\cline{2-5}
    & GPT4o-Labels & PLM-R  & 79.31\% & 79.53\%  \\\cline{2-5}
    & ---          & DeepSeekV3   & 81.23\% & 71.20\%  \\\cline{2-5}
    & DeepSeekV3-Labels & PLM-R  & 80.43\% & 79.78\% \\\hline
\multirow{4}{*}{CLASSIFIES} 
    & ---          & GPT4o        & 74.23\% & 71.52\% \\\cline{2-5}
    & GPT4o-Labels & PLM-R  & 72.79\% & 73.13\% \\\cline{2-5}
    & ---          & DeepSeekV3   & 69.08\% & 68.13\% \\\cline{2-5}
    & DeepSeekV3-Labels & PLM-R & 67.81\% & 68.13\% \\ 
\end{tabular}



    \vspace{.1in}
\caption{\small 
GPT4o and DeepSeekV3 on classification and data synthesis using the concept-based rubric in the prompt for training data labeling.
}
\label{tab:other_llm_annotation_Vs_llm}
\end{table}

In addition to GPT4o-Mini, 
we investigated the performance of GPT4o and DeepSeekV3 \cite{deepseekV3} on the classification and data synthesis.
Table \ref{tab:other_llm_annotation_Vs_llm} 
demonstrates that both two LLMs have similar performance on assessment and data synthesis.



\subsection{Explanatory Feedback} 

\begin{table}[t!]
\centering
\begin{tabular}{l|r|r|r|r}\hline
\multirow{2}{*}{\textbf{Dataset}} 
& \multicolumn{2}{|c|}{\textbf{Explainability}} & \multicolumn{2}{c}{\textbf{Subjectivity}} \\\cline{2-5}
& \multicolumn{1}{|c}{\bf Yes} & \multicolumn{1}{|c}{\bf No} & \multicolumn{1}{|c}{\bf Yes} & \multicolumn{1}{|c}{\bf No}\\\hline
ISTUDIO    & 96\%  & 4\%  & 26\% & 74\%  \\\hline
CLASSIFIES & 76\%  & 24\% & 38\% & 62\%   \\
\end{tabular}
    \vspace{.1in}
\caption{\small Feedback annotation for ISTUDIO and CLASSIFIES \textit{Partially Correct} samples with labels assigned by the LLM that are consistent with human labels.}
\label{tab:correct_prediction_feedback_annotation}
\end{table}

\begin{table}[t!]
\centering
\begin{tabular}{c|c|c|c|c|c|c}
\hline
\multirow{2}{*}{\textbf{Dataset}} & \multicolumn{2}{c|}{\textbf{Explainability}} & \multicolumn{2}{c|}{\textbf{Preference}} & \multicolumn{2}{c}{\textbf{Subjectivity}} \\ \cline{2-7}
& \textbf{ Yes } & \textbf{ No } & \textbf{LLM Score} & \textbf{Human Score} & \textbf{ Yes } & \textbf{ No } \\ \hline
ISTUDIO                            & 42\%       & 58\%      & 24\%             & 76\%              & 46\%        & 54\%        \\ \hline
CLASSIFIES                         & 16\%       & 84\%      & 10\%             & 90\%              & 76\%        & 24\%        \\ 
\end{tabular}
    \vspace{.1in}
\caption{\small Feedback annotation for ISTUDIO and CLASSIFIES samples where the LLM and human annotators disagreed.}
\label{tab:incorrect_prediction_feedback_annotation}
\end{table}

Tables \ref{tab:correct_prediction_feedback_annotation} and \ref{tab:incorrect_prediction_feedback_annotation} present the annotation results for \textbf{Explainability}, \textbf{Preference}, and \textbf{Subjectivity}, based on the three metrics outlined above (section~\ref{subsec:LLM-methods}). The \textbf{Explainability} results show that when the LLM-assigned score matches the human score, most explanations correctly align with the rubric, achieving 96\% for ISTUDIO and 76\% for CLASSIFIES (Table \ref{tab:correct_prediction_feedback_annotation}). Even when the LLM-assigned score is inconsistent with the human score, some explanations still make sense based on the rubric (Table \ref{tab:incorrect_prediction_feedback_annotation}). This partially explains why the LLM's score is sometimes preferred over the human score for student answers (as indicated in \textbf{Preference}). \textbf{This demonstrates the strong potential of using LLMs to generate feedback that explains score assignments.}

Regarding \textbf{Subjectivity}, when the LLM's score is inconsistent with the human score (Table \ref{tab:incorrect_prediction_feedback_annotation}), its explanations seem to rely more on subjective factors, such as 
clarity or communication skills, compared to cases where the LLM's score aligns with the human score (Table \ref{tab:correct_prediction_feedback_annotation}). This occurs despite the instruction in the prompt explicitly asking it to focus on the student's understanding rather than the precision of their expression (see the prompt template in Figure \ref{fig:GPT-evaluation-template}). This indicates that when assessing a student's answer involves subjective factors, the rate of disagreement between the LLM and human evaluators tends to increase.

\section{Conclusion}
The potential advantages of LLMs for formative assessment include forgoing the need for expensive training data or domain adaptation, and the potential for explanatory feedback. Yet to date, their performance has been well below supervised methods. Our results show that in-context learning (ICL) for LLMs with concept-based rubrics far surpasses example-based prompts, and narrows the assessment gap. Also, our ICL method yields LLM-generated data for training lightweight classifiers with comparable scoring accuracy with little or no human data effort.

\bibliographystyle{splncs04}
\bibliography{mybibliography}
%

\end{document}